\documentclass[10pt,twocolumn,letterpaper]{article}

\usepackage[pagenumbers]{cvpr} 
\definecolor{cvprblue}{rgb}{0.21,0.49,0.74}
\usepackage[pagebackref,breaklinks,colorlinks,allcolors=cvprblue]{hyperref}
\usepackage{multirow}
\usepackage{makecell}
\usepackage[table]{xcolor}
\def\paperID{xxx} 
\def\confName{CVPR}
\def\confYear{2026}

\title{OMGSR: You Only Need One Mid-timestep Guidance for \\ Real-World Image Super-Resolution
}
\author{Zhiqiang Wu$^{1,2}$\thanks{Work done during internship at vivo. $^\dagger$Corresponding author.} \quad Zhaomang Sun$^{2}$ \quad Tong Zhou$^{2}$ \quad Bingtao Fu$^{2}$ \quad Ji Cong$^{2}$ \\ Yitong Dong$^{2}$ \quad Huaqi Zhang$^{2}$ \quad Xuan Tang$^{1}$ \quad Mingsong Chen$^{1}$ \quad Xian Wei$^{1\dagger}$
\\
$^{1}$East China Normal University \quad
$^{2}$vivo Mobile Communication Co., Ltd.
}

\begin{document}
\maketitle
\begin{abstract}
Denoising Diffusion Probabilistic Models (DDPMs) show promising potential in one-step Real-World Image Super-Resolution (Real-ISR). Current one-step Real-ISR methods typically inject the low-quality (LQ) image latent representation at the start or end timestep of the DDPM scheduler. Recent studies have begun to note that the LQ image latent and the pre-trained noisy latent representations are intuitively closer at a mid-timestep. However, a quantitative analysis of these latent representations remains lacking. Considering these latent representations can be decomposed into signal and noise, we propose a method based on the Signal-to-Noise Ratio (SNR) to pre-compute an average optimal mid-timestep for injection. To better approximate the pre-trained noisy latent representation, we further introduce the Latent Representation Refinement (LRR) loss via a LoRA-enhanced VAE encoder. We also fine-tune the backbone of the DDPM-based generative model using LoRA to perform one-step denoising at the average optimal mid-timestep. Based on these components, we present OMGSR, a GAN-based Real-ISR framework that employs a DDPM-based generative model as the generator and a DINOv3-ConvNeXt model with multi-level discriminator heads as the discriminator. We also propose the DINOv3-ConvNeXt DISTS (Dv3CD) loss, which is enhanced for structural perception at varying resolutions. Within the OMGSR framework, we develop OMGSR-S based on SD2.1-base. An ablation study confirms that our pre-computation strategy and LRR loss significantly improve the baseline. Comparative studies demonstrate that OMGSR-S achieves state-of-the-art performance across multiple metrics. Code is available at \hyperlink{Github}{https://github.com/wuer5/OMGSR}.
\end{abstract}    
\section{Introduction}
\begin{figure}
    \centering
    \includegraphics[width=0.94\linewidth]{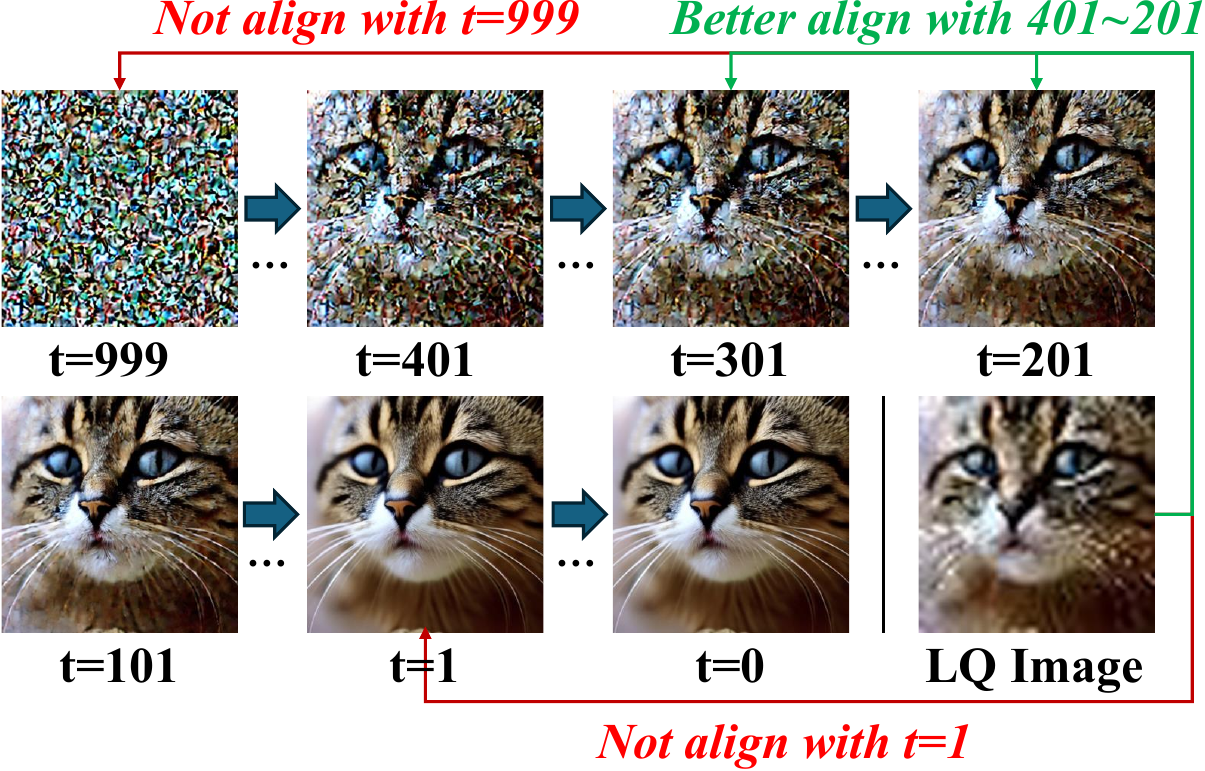}
    \caption{Illustration of a 20-step inference process with the prompt (\ie a cute cat) in SD2.1-base DDPM. We decode the latent representations into images at each step. The LQ image is sampled from the RealESRGAN degradation pipeline.}
    \label{fig:intro}
\end{figure}

Image Super-Resolution (ISR)~\cite{moser2024diffusion, wang2020deep, yang2019deep} aims to reconstruct High-Quality (HQ) images from Low-Quality (LQ) inputs, with applications in image and face restoration~\cite{imres1,face1,face_zhou}. 
Recent work focuses on Real-World ISR (Real-ISR)~\cite{cai2019toward,ren2020real}, where degradation patterns are complex and unknown. While training ISR models requires paired LQ-HQ data, collecting real-world LQ degraded images is challenging. 
Thus, synthetic degradation pipelines such as BSRGAN~\cite{bsrgan} and Real-ESRGAN~\cite{esrgan} are widely used to approximate realistic degradation.

However, synthetic LQ images often deviate from real-world representations, limiting the generalization ability of models trained on such data. Traditional supervised methods~\cite{SRCNN, ESPCN, SRGAN} perform well on synthetic pairs but struggle with real-world images due to insufficient priors~\cite{stablesr}.

Recent advances widely leverage large-scale pre-trained generative models (\eg Stable Diffusion (SD)~\cite{sd}) to fit the Real-ISR tasks due to their rich image priors~\cite{wang2024exploiting, kim2025exploiting}. Models like StableSR~\cite{stablesr}, DiffBIR~\cite{diffbir} exploit the rich priors of these foundation models, which offer stronger generalization on Real-ISR tasks and stable training via efficient fine-tuning techniques such as LoRA~\cite{lora} and ControlNet~\cite{controlnet}. 

Despite their success, these methods suffer from slow inference due to the multi-step denoising process. Recent one-step models, such as SinSR~\cite{sinsr} and OSEDiff~\cite{osediff}, significantly reduce inference time by distilling knowledge into a one step. These methods commonly encode LQ images into their latent representations, process them with a UNet~\cite{unet} for one-step denoising, and decode the output with reconstruction losses~\cite{recloss} or other knowledge distillation techniques~\cite{hinton2015distilling}.

Although one-step methods have made notable advancements, we observe that existing models (\eg OSEDiff~\cite{osediff} and PiSA-SR~\cite{pisasr}) typically inject the LQ image latent representation into the UNet in an SD model at the start ($t=999$) or end ($t=1$) timestep, creating a latent representation gap.
As illustrated in~\cref{fig:intro}, we decode the latent representations into the corresponding images in a 20-step inference process for SD2.1-base~\cite{sd}. The figure shows that LQ images align significantly more closely with images at the mid-timestep than with those at the start or end timestep. \textbf{This finding suggests that taking the LQ image latent representation as the input at a mid-timestep would narrow the latent representation gap.}

The most recent work, InvSR~\cite{invsr}, has considered this gap and set a series of mid-timesteps (\eg 250, 200, 150, 100, and 50) based on experience. However, he neither quantitatively analyzed the mid-timesteps from the perspective of pre-training nor further narrowed this gap from the parameter training standpoint.

Based on this observation and motivation, this work focuses on narrowing the gap between the LQ image latent representations and the pre-trained noisy latent representations. \textbf{We make the following contributions:}
\begin{itemize}
\item We introduce the Signal-to-Noise Ratio (SNR) based method to pre-compute an optimal mid-timestep for injection in DDPM-based generative models.
\item To further reduce the latent representation gap, we propose the Latent Representation Refinement (LRR) loss, implemented via fine-tuning the VAE encoder with LoRA to approach the pre-trained noisy latent representation.
\item We present OMGSR, a universal GAN-based Real-ISR framework compatible with DDPM-based generative models. In this framework, a generative model (LoRA-trainable) serves as the generator, while a DINOv3-ConvNeXt (frozen) with a multi-level discriminator (fully trainable) constitutes the discriminator.
\item We propose the DINOv3-ConvNeXt DISTS (Dv3CD) loss to enhance structural perception, addressing the resolution mismatch between a model's native resolution and the Real-ISR task's requirements, a discrepancy that often leads to artifacts. The Dv3CD loss natively supports training at high resolutions, such as 512, 1K, and beyond.
\item Under the OMGSR framework, we develop OMGSR-S (SD2.1-base), which achieves state-of-the-art performance across multiple benchmark metrics.
\end{itemize}

\section{Related Work}
\subsection{Multi-step Real-ISR Models} 
Recent studies have employed advanced pre-trained text-to-image (T2I) models, such as SD-based generative models~\cite{sd}, to address challenges in Real-ISR. StableSR~\cite{stablesr} leverages generative priors from SD2.1-base for Real-ISR, using a time-aware encoder and controllable feature wrapping to restore HQ images. SeeSR~\cite{seesr} utilizes high-quality semantic prompts to enhance the generative capacity of the SD2-base. 
DiffBIR~\cite{diffbir} proposes IRControlNet, which leverages the image priors of the SD2.1-base model for Real-ISR. Other multi-step Real-ISR models, such as PASD~\cite{pasd}, ResShift~\cite{resshift}, and SUPIR~\cite{superir}, also leverage powerful SD priors to achieve excellent results.
\subsection{One-step Real-ISR Models} 
Although multi-step Real-ISR models have achieved considerable success, their major drawback is their slow inference speed. Generally, an $N$-step model requires approximately $N$ times the inference time of a one-step model, a critical limitation. In addition, multi-step models tend to be more generative, and their fidelity in Real-ISR tasks is also a challenge.

Thus, one-step models have become a primary research focus in recent years due to the fast speed and high fidelity, making them suitable for Real-ISR tasks. OSEDiff~\cite{osediff} enables one-step Real-ISR by directly diffusing from the LQ image, eliminating random noise uncertainty while maintaining high-quality output through variational score distillation to conduct KL-divergence regularization.
SinSR~\cite{sinsr} achieves one-step Real-ISR by distilling a multi-step diffusion model into a student network through deterministic mapping and a novel consistency-preserving loss. PiSA-SR~\cite{pisasr} introduces a decoupled LoRA module for pixel-accurate and semantic-aware Real-ISR, enabling one-step, adjustable quality-fidelity trade-offs without retraining.

\subsection{Timestep for Injection}
For multi-step models, the majority perform multi-step denoising starting from the initial timestep (\ie $999$), following a similar training and inference process as the original SD.
In contrast, one-step models typically inject the LQ image latent representation into the UNet in an SD model at specific timesteps. For instance, OSEDiff~\cite{osediff} injects at timestep $999$, while PiSA-SR~\cite{pisasr} injects at timestep $1$. However, in pre-training, timestep $999$ corresponds to Gaussian noise, whereas timestep $1$ corresponds to a clean image, leading to a latent representation gap. In fact, the pre-trained noisy latent representation at mid-timesteps in SD is closer to the LQ image latent representation.

The most recent work, InvSR~\cite{invsr}, has considered this by empirically selecting a series of mid-timesteps for injecting the LQ image latent representation, thereby reducing the latent representation gap. Moreover, InvSR supports both one-step and multi-step inference by choosing one or multiple mid-timesteps.

Nevertheless, the selection of mid-timesteps in InvSR relies on empirical manual settings, without quantifying the selection process or further comparing the pre-trained noisy latent representation from a parameter-optimization perspective. To address this limitation, this paper explores this direction, aiming to bring the LQ image latent representation closer to the pre-trained noisy latent representation.
\section{Methodology}
In this section, we define: $\mathbf{x}_L$ and $\mathbf{x}_H$ as the LQ
image and HQ image, $\mathbf{z}_L$ and $\mathbf{z}_H$ as their corresponding
latent representations, $\mathbf{z}_t$ as the pre-trained noisy latent representation at timestep $t$, where $t\in\{1,\cdots,T\}$, $\mathcal{E}$ and $\mathcal{D}$ are VAE encoder and decoder, respectively.
\begin{figure*}
    \centering
    \includegraphics[width=0.96\linewidth]{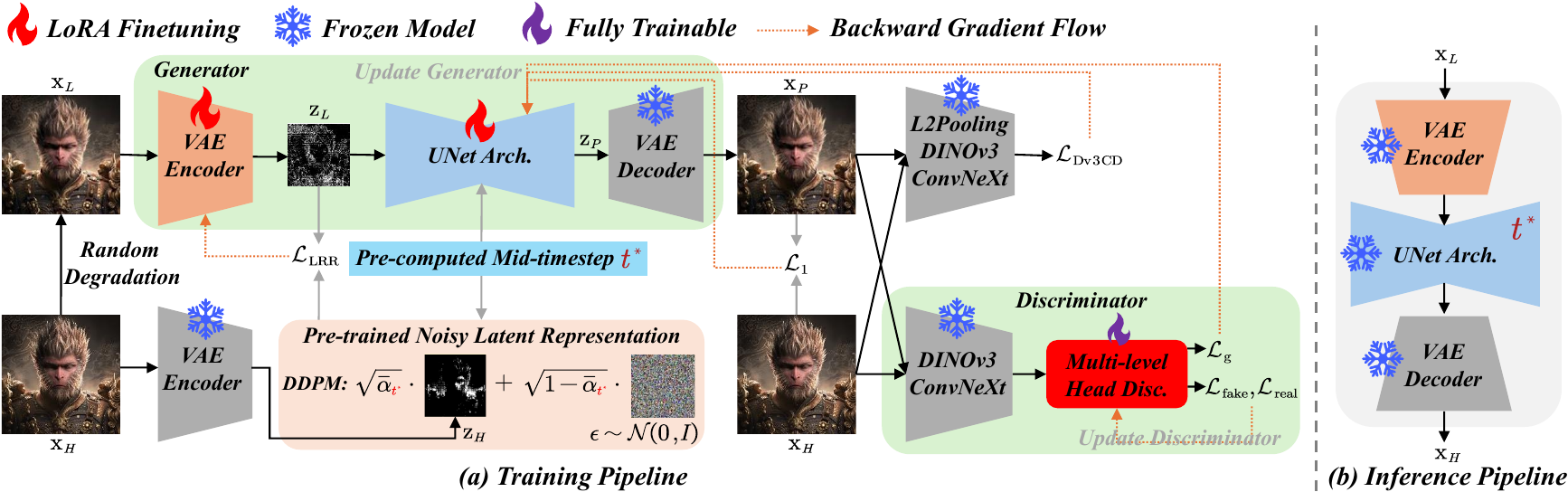}
    \caption{Illustration of our OMGSR-S training and inference pipelines. OMGSR-S achieves the fastest inference speed within SD-based Real-ISR models, as the input needs to pass through the VAE encoder, one-step prediction, and VAE decoder \textbf{only once}.}
    \label{fig:arch}
\end{figure*}

\subsection{Preliminary}
\textbf{Denoising Diffusion Probabilistic Models (DDPM~\cite{ddpm}).}
Define variances $\{\beta_1, \beta_2, \cdots, \beta_T\}$, $\alpha_t=\prod_{s=1}^t (1-\beta_s)$, the pre-trained noisy latent representation $\mathbf{z}_t$ can be combined linearly by HQ image latent representation $\mathbf{z}_H$ and Gaussian noise $\epsilon \sim \mathcal{N}(0,I)$ via a noise-added process:
\begin{equation}
\mathbf{z}_t
= \sqrt{\bar{\alpha}_t} \cdot \mathbf{z}_H + \sqrt{1-\bar{\alpha}_t} \cdot \epsilon
\label{eq1},
\end{equation}
where $\mathbf{z}_{999} \approx \epsilon$ and $\mathbf{z}_{1} \approx \mathbf{z}_H$.
In pre-training, the DDPM-based models (\eg SD2.1-base) typically use $T=999$. The pre-trained noisy latent representation $\mathbf{z}_t$, timestep $t$, and conditions $c$ (\eg prompt embeddings) are input into the UNet $\epsilon_\theta$ to predict a noise. The DDPM  training objective loss is
\begin{equation}
\mathcal{L}_{\operatorname{DDPM}}=\left\|\epsilon_\theta(\mathbf{z}_t, t, c)-\epsilon\right\|^2_2.
\end{equation}

\subsection{Motivation: Latent Representation Gap} 
One-step SD-based Real-ISR models (\eg OSEDiff~\cite{osediff} or PiSA-SR~\cite{pisasr}) fine-tune the UNet~\cite{unet} in SD2.1-base by directly feeding the LQ image latent representation $\mathbf{z}_L$ at initial timestep $999$ or end timestep $1$ \ie they assume that
$\mathbf{z}_{999} = \mathbf{z}_{L}$ or $\mathbf{z}_{1} = \mathbf{z}_{L}$. However, such assumptions create a fundamental gap since $\mathbf{z}_{999} = \epsilon$ or $\mathbf{z}_{1} = \mathbf{z}_H$ according to Eq.~\ref{eq1}. In fact, $\mathbf{z}_{L} \ne \epsilon$ or $\mathbf{z}_{L} \ne \mathbf{z}_H$ in Real-ISR tasks. 

Recent studies, such as InvSR~\cite{invsr}, have noted this problem, but they fail to quantify this gap and instead roughly select a series of mid-timesteps based on experience. We focus on quantifying this selection process and further narrowing the latent representation gap.

\subsection{Pre-computed Mid-timestep}
\subsubsection{Signal-to-Noise Ratio (SNR)}
DDPM-based models involve a pre-trained noisy latent representation $\mathbf{z}_t$ at timestep $t$, which is a linear combination of Gaussian noise $\epsilon$ and the HQ image latent representation $\mathbf{z}_H$. 
The SNR of $\mathbf{z}_t$ is:
\begin{equation}
    \texttt{SNR}(\mathbf{z}_t)=\frac{\bar{\alpha}_t  \cdot \mathbb{E}[\mathbf{z}_{H}^2]}{(1 - \bar{\alpha}_t)  \cdot\mathbb{E}[\epsilon^2]}=\frac{\bar{\alpha}_t \cdot \mathbb{E}[\mathbf{z}_H^2]}{1 - \bar{\alpha}_t}
\end{equation}
where $\epsilon \sim \mathcal{N}(0, I)$ and $\mathbb{E}[\epsilon^2]=1$.

Similarly, LQ images are obtained by applying a series of degradation processes to HQ images, where $\mathbf{z}_L$ can also be regarded as linear combinations of noise $\mathbf{z}_L - \mathbf{z}_H$ and HQ image latent representation $\mathbf{z}_H$. 
The SNR of $\mathbf{z}_L$ is
\begin{equation}
    \texttt{SNR}(\mathbf{z}_L) = \frac{\mathbb{E}[\mathbf{z}_H^2]}{\mathbb{E}[(\mathbf{z}_L - \mathbf{z}_H)^2]}
\end{equation}

\subsubsection{Average Optimal Mid-timestep}
\label{sec:332}
Given $N$ LQ-HQ image latent representation pairs $\{(\mathbf{z}_L^{(i)}, \mathbf{z}_H^{(i)})\}, i \in \{1,\cdots, N\}$ that are encoded by the VAE encoder in a dataset, we obtain an average optimal mid-timestep $t^\ast$ by traversing all $t \in \{999, \cdots, 1\}$ via:
\begin{equation}
    t^\ast = \arg \min_t \frac{1}{N}\sum_i^N \left|\texttt{SNR}(\mathbf{z}_t^{(i)}) -\texttt{SNR}(\mathbf{z}_L^{(i)})\right|
    \label{eq5}
\end{equation}

\subsection{Architecture of OMGSR}
The OMGSR framework in \cref{fig:arch}, is built on the GAN training~\cite{gan}, consisting of a generator and a discriminator.

\subsection{Generator}
Upon selecting an average optimal timestep $t^\ast$, we employ a classical DDPM-based generative model (\eg SD2.1-base~\cite{sd}). We fine-tune its VAE encoder and UNet via LoRA~\cite{lora} to perform one-step denoising by injecting the LQ image latent representation at timestep $t^\ast$, and subsequently reconstruct the pixel-level image via the VAE decoder. This process is subsequently combined with GAN-based training to accomplish the Real-ISR task. Note that, within this framework, the DDPM-based generative model is regarded as the GAN generator.

\subsubsection{Latent Representation Refinement (LRR) Loss}
Although an average optimal $t^\ast$ has been determined, we still aim to achieve sample-level refinement of the mapping from $\mathbf{z}_L$ to $\mathbf{z}_t$ through parameter fine-tuning during training. Specifically, we apply the LoRA injections to the VAE encoder $\mathcal{E}_{\varpi}$, to optimize the LoRA parameters $\varpi$. To enable this sample-wise approximation, we propose the Latent Representation Refinement (LRR) Loss:
\begin{equation}
    \begin{split}
    & \mathcal{L}_{\operatorname{LRR}}(\mathbf{x}_L, \mathbf{x}_H, t^\ast) = \mathbb{E}\left[\left\|\mathbf{z}_L - \mathbf{z}_{t^\ast}\right\|^2_2\right],
    \\
    & \mathbf{z}_L = \mathcal{E}_{\varpi}(\mathbf{x}_L),~\mathbf{z}_{t^\ast} = \sqrt{\bar{\alpha}_{t^\ast}} \cdot \mathcal{E}(\mathbf{x}_H) + \sqrt{1-\bar{\alpha}_{t^\ast}}  \cdot \epsilon.
    \end{split}
\end{equation}
Note that we adopt the L2 loss for consistency with the pre-training objective, as the KL divergence is equivalent to an L2 loss in DDPM training~\cite{ddpm}. Moreover, the L2 loss provides a more stable gradient signal during training.

\subsubsection{One-step Prediction}
After encoding via $\mathcal{E}_{\varpi}$, $\mathbf{z}_L$ becomes more closely aligned with $\mathbf{z}_{t^\ast}$. We also apply LoRA fine-tuning to the denoising UNet $\epsilon_{\theta}$. The predicted noise is then used to perform a one-step denoising process ($t^\ast \rightarrow 0$), formulated as:
\begin{equation}
    \mathbf{z}_P = \frac{\mathbf{z}_L - \sqrt{1-\bar{\alpha}_{t^\ast}}  \cdot \epsilon_\theta(\mathbf{z}_L, t^\ast, c)}{\sqrt{\bar{\alpha}_{t^\ast}}}.
    \label{eq7}
\end{equation}
Then, we obtain the predicted image: $\mathbf{x}_P = \mathcal{D}(\mathbf{z}_P)$.
\subsection{Discriminator}
\begin{figure}
    \centering
    \includegraphics[width=1\linewidth]{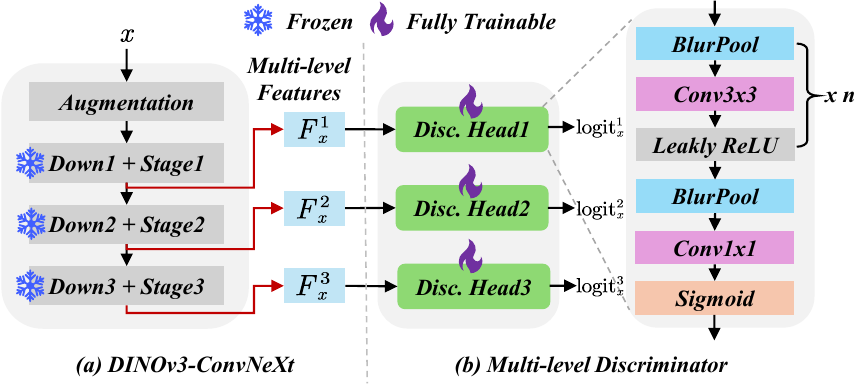}
    \caption{\textbf{(a) DINOv3-ConvNeXt} is used to extract multi-level features, which are fed into \textbf{(b) Multi-level Discriminator} to obtain the logits for discrimination. Note that $\operatorname{BlurPool}$~\cite{blurpool} is the low-pass filter used for anti-aliasing to prevent the artifacts.} 
    \label{fig:disc}
\end{figure}

\subsubsection{DINOv3-ConvNeXt Features}
\label{sec:dv3c}
The generator (\ie DDPM-based generative model) possesses powerful image priors and demonstrates exceptional image generation capabilities. Unlike traditional GANs~\cite{gan} that generate images from Gaussian noise, directly training a discriminator from scratch would lead to a significant imbalance, where the generator becomes too strong and the discriminator too weak. 

Thus, we employ DINOv3-ConvNeXt~\cite{dinov3}, which also incorporates rich image priors. We consider it for feature extraction for one main reason: its powerful extrapolation capability (achieved via RoPE~\cite{rope}), which enables seamless adaptation to Real-ISR tasks at varying resolutions (up to 4K-resolution). This is particularly important as our proposed OMGSR is a framework designed to accommodate different DDMP-based generative models across varying resolutions (\eg 1K-resolution).

Given the input image $x$, we adapt DINOv3-ConvNeXt ($\operatorname{Dv3C}$) to extract multi-level features ($l=1,2,3$): 
\begin{equation}
    \begin{split}
    &\operatorname{Dv3C}(x)=\{F^{l}_x\},~F^0_x = x,
    \\
    &F^{l}_x = \operatorname{Stage}_l(\operatorname{Down}_l(F^{l-1}_x)),
    \label{eq8}
    \end{split}
\end{equation}
where $\operatorname{Down}_l$ and $\operatorname{Stage}_l$ denote the $l$-th downsampling and feature extraction layers in DINOv3-ConvNeXt, respectively. The model comprises four main stages, enabling feature extraction at four distinct levels. The first three levels primarily contain detailed information, whereas the final level mainly encompasses global semantic information. Since $\mathbf{z}_P$ in \cref{eq7} is denoised from $\mathbf{z}_L$ and already incorporates global semantic information but lacks detailed information, only the first three levels are utilized for feature extraction.
\subsubsection{Multi-level Discriminator}

We design the multi-level discriminator heads $\operatorname{DiscHead}_l$, which take $\{F^{l}_x\},~l=1,2,3$ as the input to predict the multi-level logits for discrimination, as illustrated in \cref{fig:disc}. We define the discrimination process as:
\begin{equation}
    \begin{split}
        &\operatorname{Disc}(x) = \{\operatorname{logit}_x^l\},
        \\
        &\operatorname{logit}_x^l = \operatorname{Sigmoid}(\operatorname{DiscHead}_l(F^{l}_x)).
    \end{split}
\end{equation}

\subsection{Generator Losses}
\subsubsection{DINOv3-ConvNeXt DISTS (Dv3CD) Loss}
In GAN training~\cite{gan}, structural perception loss has been shown to play a critical role by preserving global structural information of the image, while the GAN component focuses on refining local details. Most existing methods adopt LPIPS~\cite{lpips} as the perceptual loss. However, we note that LPIPS is built on a VGG~\cite{vgg} pre-trained on $224\times224$ images, which limits its effectiveness when training GANs at higher resolutions such as $512$ or 1K, often leading to structural artifacts in the generated images. 

We propose the DINOv3-ConvNeXt DISTS (Dv3CD) loss to address the resolution-mismatch problem. Following \cref{eq8}, we also utilize DINOv3-ConvNeXt and incorporate an $\operatorname{L2Pooling}_l$ after each $l$-th stage to gather global structural information for structural perception. The main reason we use DINOv3-ConvNeXt is the same as in \cref{sec:dv3c}: its powerful extrapolation capability.

Given the input image $x$, we define the L2-pooling DINOv3-ConvNeXt ($\operatorname{L2Dv3C}$) to extract the first three level L2-pooling features ($l=1,2,3$):
\begin{equation}
    \begin{split}
    &\operatorname{L2Dv3C}(x)=\{L^{l}_x\},~L^0_x=x,
    \\
    &L^{l}_x = \operatorname{L2Pooling}_l(\operatorname{Stage}_l(\operatorname{Down}_l(L^{l-1}_x))),\\
    &\operatorname{L2Pooling}(\Box) = \sqrt{\operatorname{Conv2d}(\Box^2, w_h)}, 
    \end{split}
\end{equation}
where $w_h$ is the 2D Hanning window filters~\cite{hann}.

Then we obatin the L2-pooling features of $\mathbf{x}_P,~\mathbf{x}_H$ via $\operatorname{L2Dv3C}(\mathbf{x}_P)=\{L^{l}_{\mathbf{x}_P}\},~\operatorname{L2Dv3C}(\mathbf{x}_H)=\{L^{l}_{\mathbf{x}_H}\}$ in our OMGSR framework. We also include $L^0_{\mathbf{x}_P}=\mathbf{x}_P$, $L^0_{\mathbf{x}_H}=\mathbf{x}_H$ for caculation.
Then Dv3CD loss is caculated via the DISTS~\cite{dists} ($l=0,1,2,3$): 
\begin{equation}
\begin{split}
    &\mathcal{L}_{\operatorname{Dv3CD}}(\mathbf{x}_P, \mathbf{x}_H) = \operatorname{DISTS}(
    \{L^{l}_{\mathbf{x}_P}\}, \{L^{l}_{\mathbf{x}_H}\}
    )
    \\
    &=1 - \frac{1}{\alpha} \sum^3_{l=0} \mathbb{E}\left[ \frac{2\mu_{L^l_{\mathbf{x}_P}} \mu_{L^l_{\mathbf{x}_H}}}{\mu_{L^l_{\mathbf{x}_P}}^2 + \mu_{L^l_{\mathbf{x}_H}}^2} + \frac{2\sigma_{L^l_{\mathbf{x}_P}, L^l_{\mathbf{x}_H}}}{\sigma_{L^l_{\mathbf{x}_P}}^2 + \sigma_{L^l_{\mathbf{x}_H}}^2}\right],
\end{split}
\end{equation}
where $\mu_\Box$, $\sigma_\Box$, and $\sigma_{\Box,\Box}$ denote the mean, variance, and covariance computed across the 2D spatial dimensions of the feature maps, respectively. The coefficient $\alpha$ is defined as $\alpha = 2\sum_{l=0}^{3} C(L_{\mathbf{x}_H}^l)$, where the function $C(\Box)$ denotes the number of channels.
\subsubsection{Generator GAN Loss}
We directly employ the standard GAN loss, as the GAN training in Real-ISR follows an LQ$\rightarrow$HQ mapping rather than a Noise$\rightarrow$HQ paradigm, making the learning task relatively easy and stable. We obtain the logits of $\mathbf{x}_P$ via $\operatorname{Disc}(\mathbf{x}_P) = \{\operatorname{logit}^l_{\mathbf{x}_P}\}$. Then the generator GAN loss is:
\begin{equation}
\begin{split}
    &\mathcal{L}_{\operatorname{g}}(\mathbf{x}_P) = \sum^{3}_{l=1} \mathbb{E}\left[\operatorname{BCE}(\operatorname{logit}^l_{\mathbf{x}_P}, \operatorname{label}_{\operatorname{real}})\right],
\end{split}
\end{equation}
where $\operatorname{label}_{\operatorname{real}}$ is the soft label, and set to $0.8$. The function $\operatorname{BCE}$ denotes the binary cross-entropy.
\subsubsection{Complete Generator Loss}
For the generator, a structural perception loss is required to preserve the overall structure, while the generator GAN loss ensures the fine details. We also employ an L1 loss $\mathcal{L}_1$ for pixel-level constraints. We only update the parameters of the VAE encoder and UNet. The complete generator loss is 
\begin{equation}
    \label{eq13}
    \begin{split}
    &\mathcal{L}_{\operatorname{G}}(\mathbf{x}_L, \mathbf{x}_P, \mathbf{x}_H) 
    = \lambda_1 \mathcal{L}_{\operatorname{LRR}}(\mathbf{x}_L, \mathbf{x}_H, t^\ast) + \lambda_2 \mathcal{L}_{\operatorname{g}}(\mathbf{x}_P) 
    \\
    &+ \lambda_3 \mathcal{L}_{\operatorname{Dv3CD}}(\mathbf{x}_P, \mathbf{x}_H) + \lambda_4 \mathcal{L}_{1}(\mathbf{x}_P, \mathbf{x}_H).
    \end{split}
\end{equation}
\subsection{Discriminator Losses}
\subsubsection{Discriminator GAN Loss}
For discriminator training, we only update its parameters, aiming to discriminate between fake and real images. We obtain the logits of $\mathbf{x}_{\widehat{P}}=\operatorname{Detach}(\mathbf{x}_P),~\mathbf{x}_H$ via $\operatorname{Disc}(\mathbf{x}_{\widehat{P}}) = \{\operatorname{logit}^l_{\mathbf{x}_{\widehat{P}}}\},~\operatorname{Disc}(\mathbf{x}_H) = \{\operatorname{logit}^l_{\mathbf{x}_H}\}$. Then the discriminator GAN loss is:
\begin{equation}
\begin{split}
    &\mathcal{L}_{\operatorname{fake}}(\mathbf{x}_{\widehat{P}}) = \sum^{3}_{l=1} \mathbb{E}\left[\operatorname{BCE}(\operatorname{logit}^l_{\mathbf{x}_{\widehat{P}}}, \operatorname{label}_{\operatorname{fake}})\right], 
    \\
    &\mathcal{L}_{\operatorname{real}}(\mathbf{x}_H) = \sum^{3}_{l=1} \mathbb{E}\left[\operatorname{BCE}(\operatorname{logit}^l_{\mathbf{x}_H}, \operatorname{label}_{\operatorname{real}})\right],
\end{split}
\end{equation}
where $\operatorname{label}_{\operatorname{fake}}$ is set to $0$.
\subsubsection{Complete Discriminator Loss}
The complete discriminator loss consists of $\mathcal{L}_{\operatorname{fake}}$ and $\mathcal{L}_{\operatorname{real}}$:
\begin{equation}
    \mathcal{L}_{\operatorname{D}}(\mathbf{x}_{\widehat{P}}, \mathbf{x}_H) = \lambda_2 (\mathcal{L}_{\operatorname{fake}}(\mathbf{x}_{\widehat{P}}) + \mathcal{L}_{\operatorname{real}}(\mathbf{x}_H)).
\end{equation}

\section{Experiments}
\subsection{Experimental Settings}
\subsubsection{Training Settings.} We follow SeeSR~\cite{seesr}, utilizing LSDIR~\cite{lsdir} dataset and the first 10,000 face images from FFHQ~\cite{ffhq} for the $\times 4$ SR task. LQ-HQ image pairs are synthesized via the standard Real-ESRGAN~\cite{esrgan} degradation pipeline. Within the OMGSR framework, we adapt the SD2.1 base as the foundation model to instantiate OMGSR-S. For LoRA injection, we set the VAE LoRA
rank to $16$ and the UNet LoRA rank to $32$. We employ the AdamW~\cite{adamw} optimizer with a learning rate of $5 \times 10^{-5}$, a batch size of $1$, and $4$ gradient steps, training on four RTX 4090 GPUs for $5100$ steps. We set $\lambda_1=5, \lambda_2=0.5, \lambda_3=5, \lambda_1=0.5$ in \cref{eq13}.

\subsubsection{Test Dataset} Since OMGSR focuses on Real-ISR tasks, we test it on four classic datasets: RealSR~\cite{realsr}, DrealSR~\cite{drealsr}, RealLQ250~\cite{reallq250}, and RealLR200~\cite{seesr}. Among these, RealSR contains $100$ images of size $128 \times 128$, and DrealSR includes $93$ images of the same size. RealLQ250 comprises $250$ images with dimensions $256 \times 256$, while RealLR200 consists of $200$ images with varying resolutions. For RealLQ250 and RealLR200, tiled diffusion is required during testing. We conduct all experiments on the $\times4$ Real-ISR task.

\subsubsection{Metrics} Traditional reference metrics like PSNR, SSIM~\cite{ssim} struggle to reflect the real visual effects of the generated images~\cite{blau2018perception, jinjin2020pipal, yu2024scaling}. Thus, we use model-based reference metrics: LPIPS~\cite{lpips} and FID~\cite{fid}; non-reference metrics: CLIPIQA~\cite{clipiqa}, CLIPIQA+~\cite{clipiqa}, NIMA~\cite{nima}, NIQE~\cite{niqe}, LIQE~\cite{liqe}, MUSIQ~\cite{musiq}, and MANIQA~\cite{maniqa}. These metrics reflect the image quality in multiple ways. In this paper, we use the PyIQA~\cite{pyiqa} tool to evaluate these metrics.

\subsection{Ablation Study}
\subsubsection{Ablation Study on Timestep}
\label{sec:421}
\begin{figure}
    \centering
    \includegraphics[width=1\linewidth]{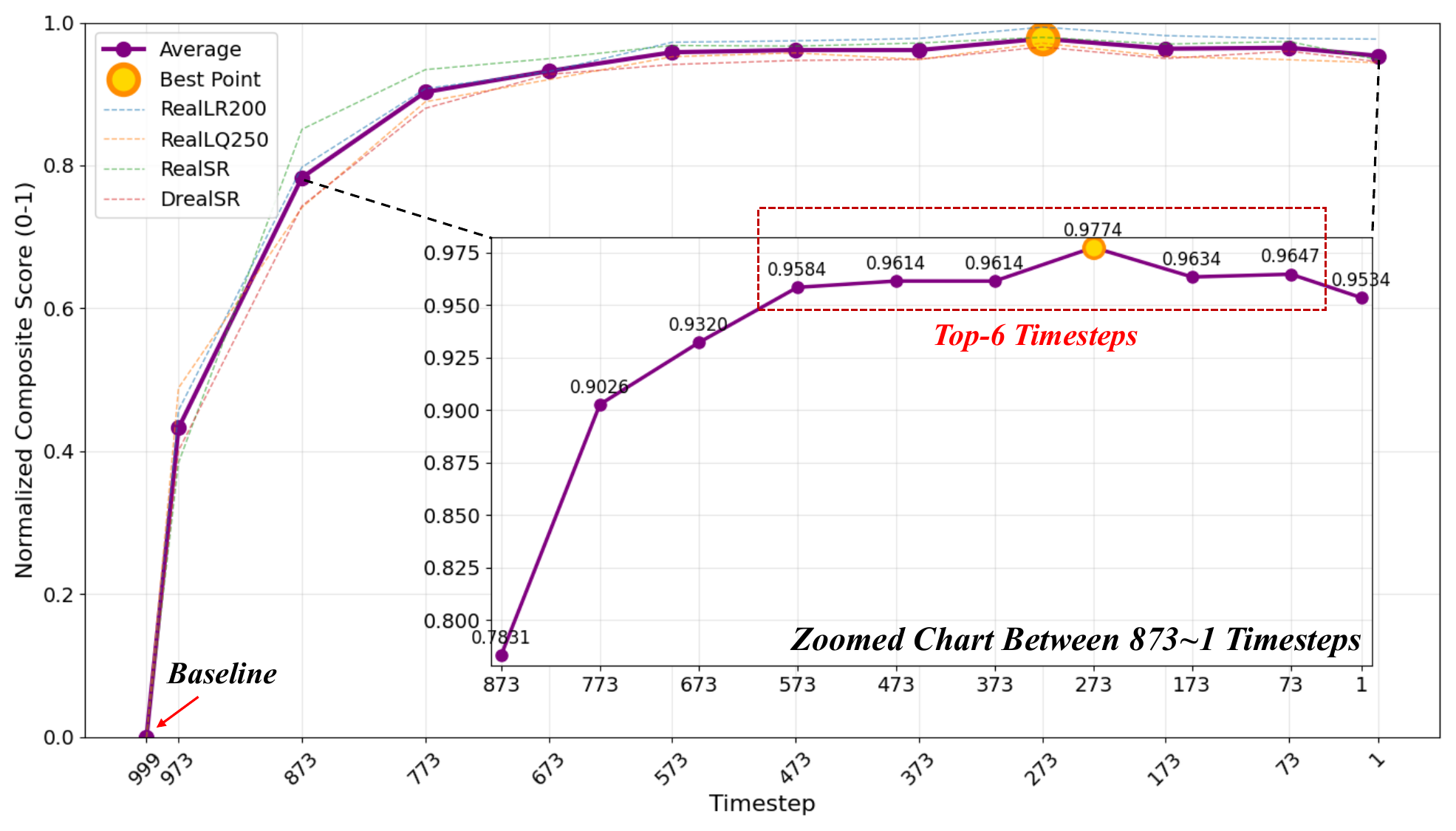}
    \caption{Ablation Study of OMGSR-S with different timesteps. We set the timestep $273$ as the baseline, with intervals of $100$. Note that we conduct experiments on OMGSR-S \textbf{without} the proposed LRR Loss to avoid any impact on the latent representation. We report a normalized score based on $9$ metrics across four datasets. The details are in the \textbf{supplementary materials}.}
    \label{fig:ablation_timestep}
\end{figure}

We ablate the different timesteps using OMGSR-S \textbf{without} LRR loss. Note that we do not include LRR loss to avoid its effect on latent representation. The baseline timestep is $273$ calculated from ~\cref{eq5}. Performance is evaluated at 100-timestep intervals, with results aggregated into a composite score ($0$–$1$) using normalized metrics across four datasets.

As shown in \cref{fig:ablation_timestep}, injecting the LQ image latent $\mathbf{z}_L$ in early timesteps ($999$–$673$) severely degrades performance, whereas mid-to-late injections ($573$–$173$) yield high and stable results. The peak occurs at timestep $273$, aligning with the theoretically derived value in~\cref{eq5}. This confirms that our SNR-based method effectively guides the selection of timesteps in DDPM-based Real-ISR.

\subsubsection{Ablation Study on Architecture}
\begin{table}
  \caption{Ablation study of OSEDiff with $t=999$ \vs $t=273$.}
  \label{tab:osediff}
  \resizebox{\columnwidth}{!}{
  \centering
  \begin{tabular}{c|c|cccccccc}
    \toprule
    Dataset & Model & CLIPIQA$\uparrow$ & CLIPIQA+$\uparrow$ & NIMA$\uparrow$ & NIQE$\downarrow$ & LIQE$\uparrow$ & MUSIQ$\uparrow$ & MANIQA$\uparrow$ \\
    \midrule
    \multirow{2}{*}{RealSR} & 999 & {0.6582} & {0.6798} & {4.7581} & {5.4996} & {3.7536} & {67.5307} & {0.6223} \\
    & 273 & \textbf{0.6589} & \textbf{0.6853} & \textbf{4.7760} & \textbf{5.4132} & \textbf{3.8060} & \textbf{68.0663} & \textbf{0.6400} \\
    \midrule 
    \multirow{2}{*}{DrealSR} & 999 & \textbf{0.6937} & \textbf{0.6864} & \textbf{4.6248} & {6.3676} & \textbf{3.7991} & {64.1535} & {0.5917} \\
    & 273 & {0.6857} & {0.6808} & {4.6136} & \textbf{5.8439} & {3.7600} & \textbf{64.7893} & \textbf{0.6072} \\
    \midrule 
    \multirow{2}{*}{RealLR250} & 999 & {0.7103} & {0.7131} & {5.1855} & {4.0279} & {3.8323} & {70.2660} & {0.6019} \\
    & 273 & \textbf{0.7353} & \textbf{0.7349} & \textbf{5.2604} & \textbf{3.8477} & \textbf{4.0427} & \textbf{71.5111} & \textbf{0.6251} \\
    \midrule
    \multirow{2}{*}{RealLR200} & 999 & {0.7077} & {0.7154} & {5.3150} & {4.0738} & {3.9358} & {69.4883} & {0.6194} \\
    & 273 & \textbf{0.7281} & \textbf{0.7379} & \textbf{5.3817} & \textbf{3.9197} & \textbf{4.1321} & \textbf{70.9423} & \textbf{0.6421} \\
    \bottomrule
  \end{tabular}
  }
\end{table}
To further validate ~\cref{eq5}, we conduct an ablation study on OSEDiff~\cite{osediff} with $t=999$ (original paper settings) \vs $t=273$ (our paper settings). Note that the degradation configuration of OSEDiff is the same as that of OMGSR-S, thus $273$ is also available to OSEDiff. We keep all basic OSEDiff settings unchanged and train both models for 17,000 steps with a batch size of $1$ and a gradient step size of $4$ on four NVIDIA RTX 4090 GPUs.

As shown in~\cref{tab:osediff}, OSEDiff ($t=273$) significantly outperforms OSEDiff ($t=999$) across all datasets and metrics. Most metrics (\eg MANIQA, MUSIQ) of OSEDiff ($t=273$) also surpass those reported in~\cref{tab:cmp}. These results further confirm our analysis of mid-timestep and the effectiveness of our SNR-based method in~\cref{eq5}.

\subsubsection{Ablation study on LRR Loss} 
\begin{table}
  \caption{Ablation Study of $\mathcal{L}_{\operatorname{LRR}}$. The baseline denotes OMGSR-S without the $\mathcal{L}_{\operatorname{LRR}}$. We define $\mathcal{L}_{\operatorname{TOHQ}}=\mathbb{E}[\|\mathbf{z}_L-\mathbf{z}_H\|^2_2]$.}
  \label{tab:lrr}
  \resizebox{\columnwidth}{!}{
  \centering
  \begin{tabular}{c|c|cccccccc}
    \toprule
    Dataset & Method & CLIPIQA$\uparrow$ & CLIPIQA+$\uparrow$ & NIMA$\uparrow$ & NIQE$\downarrow$ & LIQE$\uparrow$ & MUSIQ$\uparrow$ & MANIQA$\uparrow$ \\
    \midrule
    \multirow{3}{*}{RealSR} & baseline & 0.6352 & 0.6785 & 4.9030 & 5.6487 & 4.1373 & 68.5648 & 0.6756 \\
    & w/ $\mathcal{L}_{\operatorname{TOHQ}}$ & \underline{0.6607} & \underline{0.7025} & \textbf{5.0079} & \underline{5.5170} & \underline{4.3073} & \underline{69.4674} & \underline{0.6701} \\
    & w/ $\mathcal{L}_{\operatorname{LRR}}$ & \textbf{0.6709} & \textbf{0.7143} & \underline{5.0025} & \textbf{5.1893} & \textbf{4.3509} & \textbf{69.9406} & \textbf{0.6753}
    \\ 
    \midrule 
    \multirow{3}{*}{DrealSR} & baseline & 0.6552 & 0.6650 & 4.7416 & 6.5184 & 3.9885 & 65.3381 & 0.6259 \\
    & w/ $\mathcal{L}_{\operatorname{TOHQ}}$ & \textbf{0.6993} & \underline{0.6962} & \underline{4.7910} & \underline{6.0305} & \underline{4.2286} & \underline{66.2147} & \underline{0.6276} \\
    & w/ $\mathcal{L}_{\operatorname{LRR}}$ & \underline{0.6958} & \textbf{0.7053} & \textbf{4.8008} & \textbf{5.7720} & \textbf{4.2775} & \textbf{66.6773} & \textbf{0.6354}
    \\
    \midrule 
    \multirow{3}{*}{RealLR250} & baseline & 0.6976 & 0.7065 & 5.1490 & 4.0240 & 3.9954 & 69.8671 & 0.6302\\
    & w/ $\mathcal{L}_{\operatorname{TOHQ}}$ & \underline{0.7176} & \underline{0.7217} & \underline{5.1664} & \underline{3.9883} & \underline{4.1499} & \underline{70.2378} & \underline{0.6287} \\
    & w/ $\mathcal{L}_{\operatorname{LRR}}$ & \textbf{0.7318} & \textbf{0.7352} & \textbf{5.2356} & \textbf{3.8771} & \textbf{4.2546} & \textbf{71.0128} & \textbf{0.6385}
    \\
    \midrule
    \multirow{3}{*}{RealLR200} & baseline & 0.7016 & 0.7211 & 5.3617 & 4.0731 & 4.1901 & 69.8960 & 0.6571\\
    & w/ $\mathcal{L}_{\operatorname{TOHQ}}$ & \underline{0.7170} & \underline{0.7332} & \underline{5.3649} & \underline{4.0693} & \underline{4.2865} & \underline{70.4493} & \underline{0.6582} \\
    & w/ $\mathcal{L}_{\operatorname{LRR}}$ & \textbf{0.7297} & \textbf{0.7456} & \textbf{5.3928} & \textbf{3.9628} & \textbf{4.3806} & \textbf{71.0941} & \textbf{0.6646}
    \\
\bottomrule
  \end{tabular}
  }
\end{table}

\begin{table*}[h]
  \caption{Quantitative comparisons ($\times 4$) with state-of-the-art multi-step and one-step models on four real-world benchmark datasets. The best and second-best results of each metric are highlighted in {bold} and {underlined}, respectively. "-s" denotes the number of diffusion steps.}
  \label{tab:cmp}
  \resizebox{\textwidth}{!}{ 
  \centering
  \begin{tabular}{c|c|ccccc|cccccc}
    \toprule
    Datasets & Metrics & RealESRGAN~\cite{esrgan} & \makecell{StableSR~\cite{stablesr}\\-s200} & \makecell{SeeSR~\cite{seesr}\\-s50} & \makecell{DiffBIR~\cite{diffbir}\\-s50} & \makecell{ResShift~\cite{resshift}\\-s15} & \makecell{SinSR~\cite{sinsr}\\-s1} & \makecell{OSEDiff~\cite{osediff}\\-s1} & \makecell{InvSR~\cite{invsr}\\-s1} & \makecell{PiSA-SR~\cite{pisasr}\\-s1} & \makecell{OMGSR-S\\-s1} \\
    \midrule
    \multirow{9}{*}{RealSR} 
     & LPIPS$\downarrow$ & 0.2710 & \textbf{0.2604} & 0.3007 & 0.3470 & 0.3159 & 0.3210 & 0.2921 & 0.2871 & \underline{0.2672} & 0.3009 \\
     & FID$\downarrow$ & 135.15 & 132.09 & 125.51 & 134.59 & 149.65 & 136.78 & \textbf{123.50} & 138.85 & 124.19 & \underline{124.05} \\
     & CLIPIQA$\uparrow$ & 0.4490 & 0.5426 & 0.6699 & \textbf{0.6960} & 0.5505 & 0.6156 & 0.6693 & \textbf{0.6785} & 0.6699 & 0.6709 \\
     & CLIPIQA+$\uparrow$ & 0.5841 & 0.6150 & 0.6909 & \underline{0.6989} & 0.5451 & 0.5370 & 0.6964 & 0.6880 & 0.6957 & \textbf{0.7143} \\
     & NIMA$\uparrow$ & 4.6551 & 4.6767 & 4.9191 & 4.9159 & 4.7554 & 4.6643 & 4.8951 & \textbf{5.0946} & 4.8953 & \underline{5.0025} \\
     & NIQE$\downarrow$ & 5.7960 & 6.6231 & \underline{5.3984} & 5.4992 & 6.8833 & 6.2998 & 5.6474 & 5.6222 & 5.5057 & \textbf{5.1893} \\
     & LIQE$\uparrow$ & 3.3571 & 3.2578 & 4.1354 & 4.0261 & 3.1859 & 3.1466 & 4.0690 & 4.0392 & \underline{4.0989} & \textbf{4.3509} \\
     & MUSIQ$\uparrow$ & 60.3657 & 61.8058 & 69.8165 & 68.3462 & 60.2181 & 60.4204 & 69.0896 & 68.5372 & \textbf{70.1492} & \underline{69.9406} \\
     & MANIQA$\uparrow$ & 0.5492 & 0.5952 & 0.6445 & 0.6540 & 0.5388 & 0.5391 & 0.6331 & \underline{0.6628} & 0.6552 & \textbf{0.6753} \\
    \midrule
    \multirow{9}{*}{DrealSR} 
     & LPIPS$\downarrow$ & \underline{0.2819} & \textbf{0.2698} & 0.3174 & 0.4520 & 0.3526 & 0.3674 & 0.2968 & 0.3538 & 0.2960 & 0.3267 \\
     & FID$\downarrow$ & 147.80 & 151.27 & 147.53 & 177.06 & 176.70 & 171.88 & \underline{135.29} & 171.40 & \textbf{130.43} & 142.44 \\
     & CLIPIQA$\uparrow$ & 0.4517 & 0.4907 & 0.6913 & 0.6859 & 0.5410 & 0.6348 & 0.6963 & \textbf{0.7132} & \underline{0.6974} & 0.6958 \\
     & CLIPIQA+$\uparrow$ & 0.5544 & 0.5347 & 0.6794 & 0.6828 & 0.5157 & 0.5400 & 0.6825 & \underline{0.6832} & 0.6920 & \textbf{0.7053} \\
     & NIMA$\uparrow$ & 4.3261 & 4.2136 & 4.6945 & 4.7847 & 4.4405 & 4.4639 & 4.6766 & \textbf{4.8566} & 4.6250 & \underline{4.8008} \\
     & NIQE$\downarrow$ & 6.6927 & 7.5488 & 6.4136 & 6.2409 & 7.8693 & 7.1422 & 6.4904 & \underline{5.9917} & 6.1759 & \textbf{5.7720} \\
     & LIQE$\uparrow$ & 2.9259 & 2.4349 & \underline{4.1268} & 3.8930 & 2.7968 & 3.0514 & 3.9371 & 4.0557 & 4.0440 & \textbf{4.2775} \\
     & MUSIQ$\uparrow$ & 54.2721 & 51.3635 & 65.0935 & 65.6585 & 52.3726 & 54.9825 & 64.6537 & 65.9956 & \underline{66.1094} & \textbf{66.6773} \\
     & MANIQA$\uparrow$ & 0.4899 & 0.4969 & 0.6043 & 0.6279 & 0.4750 & 0.4855 & 0.5895 & \underline{0.6302} & 0.6146 & \textbf{0.6354} \\
    \midrule
    \multirow{7}{*}{RealLQ250} 
     & CLIPIQA$\uparrow$ & 0.5434 & 0.5150 & 0.7132 & \underline{0.7255} & 0.4734 & 0.6998 & 0.6995 & 0.6628 & 0.7095 & \textbf{0.7318} \\
     & CLIPIQA+$\uparrow$ & 0.6117 & 0.5811 & 0.7142 & \underline{0.7213} & 0.4642 & 0.5919 & 0.7017 & 0.6722 & 0.7160 & \textbf{0.7352} \\
     & NIMA$\uparrow$ & 5.2554 & 5.0700 & 5.3863 & \textbf{5.4922} & 5.0243 & 5.1938 & 5.2364 & \underline{5.4401} & 5.2429 & 5.2356 \\
     & NIQE$\downarrow$ & 4.1292 & 4.6345 & 3.9832 & \textbf{3.5608} & 4.8476 & 5.7974 & 3.9127 & 4.4098 & \underline{3.8751} & 3.8771 \\
     & LIQE$\uparrow$ & 3.3410 & 2.7533 & \underline{4.1336} & 4.0874 & 2.4609 & 3.2465 & 3.8610 & 3.7113 & 3.9813 & \textbf{4.2546} \\
     & MUSIQ$\uparrow$ & 62.5169 & 57.1341 & \textbf{71.1218} & 70.3687 & 57.7724 & 63.8548 & 69.6786 & 65.8212 & \underline{71.0710} & 71.0128 \\
     & MANIQA$\uparrow$ & 0.5288 & 0.5203 & 0.6204 & \underline{0.6232} & 0.4657 & 0.5178 & 0.5928 & 0.5914 & 0.6157 & \textbf{0.6385} \\
    \midrule
    \multirow{7}{*}{RealLQ200} 
     & CLIPIQA$\uparrow$ & 0.5409 & 0.5731 & 0.6959 & \underline{0.7222} & 0.4942 & 0.6615 & 0.7008 & 0.6830 & 0.7125 & \textbf{0.7297} \\
     & CLIPIQA+$\uparrow$ & 0.6222 & 0.6360 & 0.7171 & 0.7241 & 0.5094 & 0.5888 & 0.7136 & 0.7111 & \underline{0.7292} & \textbf{0.7456} \\
     & NIMA$\uparrow$ & 5.1866 & 5.2607 & 5.4154 & \underline{5.4648} & 5.0419 & 5.1763 & 5.3530 & \textbf{5.5045} & 5.3920 & 5.3928 \\
     & NIQE$\downarrow$ & 4.1796 & 4.3515 & 3.9996 & \textbf{3.7803} & 4.8432 & 5.3042 & \underline{3.9268} & 3.9936 & 3.9991 & 3.9628 \\
     & LIQE$\uparrow$ & 3.4836 & 3.4319 & \underline{4.1806} & 4.0541 & 2.6938 & 3.2175 & 3.9967 & 4.0778 & 4.1690 & \textbf{4.3806} \\
     & MUSIQ$\uparrow$ & 62.9605 & 63.3346 & 70.2502 & 68.7120 & 57.4580 & 61.3634 & 69.5654 & 68.9061 & \underline{70.8834} & \textbf{71.0941} \\
     & MANIQA$\uparrow$ & 0.5553 & 0.5749 & 0.6360 & 0.6385 & 0.4925 & 0.5343 & 0.6143 & 0.6481 & \underline{0.6418} & \textbf{0.6646} \\
    \bottomrule
  \end{tabular}
  }
\end{table*}

\begin{table}
  \caption{Ablation study of OMGSR-S with $\mathcal{L}_{\operatorname{LPIPS}}$ \vs $\mathcal{L}_{\operatorname{Dv3CD}}$.}
  \label{tab:Dv3CD}
  \resizebox{\columnwidth}{!}{
  \centering
  \begin{tabular}{c|c|cccccccc}
    \toprule
    Dataset & Loss & CLIPIQA$\uparrow$ & CLIPIQA+$\uparrow$ & NIMA$\uparrow$ & NIQE$\downarrow$ & LIQE$\uparrow$ & MUSIQ$\uparrow$ & MANIQA$\uparrow$ \\
    \midrule
    \multirow{2}{*}{RealSR} & $\mathcal{L}_{\operatorname{LPIPS}}$ & 0.6422 & 0.7052 & 4.8855 & 5.2712 & 4.2258 & 69.6518 & \textbf{0.6776} \\
    & $\mathcal{L}_{\operatorname{Dv3CD}}$ & \textbf{0.6709} & \textbf{0.7143} & \textbf{5.0025} & \textbf{5.1893} & \textbf{4.3509} & \textbf{69.9406} & 0.6753
    \\ 
    \midrule 
    \multirow{2}{*}{DrealSR} & $\mathcal{L}_{\operatorname{LPIPS}}$ & 0.6732 & 0.6867 & 4.6917 & 5.9989 & 4.1462 & 65.9111 & 0.6353 \\
    & $\mathcal{L}_{\operatorname{Dv3CD}}$ & \textbf{0.6958} & \textbf{0.7053} & \textbf{4.8008} & \textbf{5.7720} & \textbf{4.2775} & \textbf{66.6773} & \textbf{0.6354}
    \\
    \midrule 
    \multirow{2}{*}{RealLR250} & $\mathcal{L}_{\operatorname{LPIPS}}$ & 0.7179 & 0.7276 & 5.2164 & \textbf{3.7442} & 4.2085 & \textbf{71.3788} & \textbf{0.6415} \\
    & $\mathcal{L}_{\operatorname{Dv3CD}}$ & \textbf{0.7318} & \textbf{0.7352} & \textbf{5.2356} & 3.8771 & \textbf{4.2546} & 71.0128 & 0.6385
    \\
    \midrule
    \multirow{2}{*}{RealLR200} & $\mathcal{L}_{\operatorname{LPIPS}}$ & 0.7121 & 0.7354 & 5.3271 & \textbf{3.9039} & 4.2888 & 70.8682 & 0.6610 \\
    & $\mathcal{L}_{\operatorname{Dv3CD}}$ & \textbf{0.7297} & \textbf{0.7456} & \textbf{5.3928} & 3.9628 & \textbf{4.3806} & \textbf{71.0941} & \textbf{0.6646}
    \\
\bottomrule
  \end{tabular}
  }
\end{table}

To evaluate the effectiveness of the proposed $\mathcal{L}_{\operatorname{LRR}}$, we ablate OMGSR-S under three settings: (1) \textbf{without} $\mathcal{L}_{\operatorname{LRR}}$ (baseline), (2) \textbf{with} $\mathcal{L}_{\operatorname{LRR}}$, and (3) \textbf{with} $\mathcal{L}_{\operatorname{TOHQ}}$, a variant that approximates the HQ image latent $\mathbf{z}_H$ instead of the noisy pre-trained latent $\mathbf{z}_{t^*}$ \ie $\mathcal{L}_{\operatorname{TOHQ}} = \mathbb{E}[|\mathbf{z}_L - \mathbf{z}_H|^2_2]$.

Results in~\cref{tab:lrr} show two key findings:
\textbf{First}, both $\mathcal{L}_{\operatorname{LRR}}$ and $\mathcal{L}_{\operatorname{TOHQ}}$ bring consistent and significant gains over the baseline across all datasets and metrics, confirming that guiding $\mathbf{z}_L$ toward either $\mathbf{z}_H$ or $\mathbf{z}_{t^*}$ is beneficial.
\textbf{Second}, $\mathcal{L}_{\operatorname{LRR}}$ outperforms $\mathcal{L}_{\operatorname{TOHQ}}$ on nearly all metrics, indicating that aligning with the pre-trained noisy latent $\mathbf{z}_{t^*}$ better activates the generative prior by adhering to the model's inherent training rules.

\subsubsection{Ablation Study on Dv3CD Loss} 
We ablate two structural perception losses: $\mathcal{L}_{\operatorname{LPIPS}}$ and $\mathcal{L}_{\operatorname{Dv3CD}}$ on OMGSR-S. As shown in~\cref{tab:Dv3CD}, $\mathcal{L}_{\operatorname{Dv3CD}}$ achieves better results in $23$ out of $28$ metrics across four datasets. Although $\mathcal{L}_{\operatorname{LPIPS}}$ yields slightly better NIQE on two datasets, $\mathcal{L}_{\operatorname{Dv3CD}}$ shows a clear overall advantage.

\begin{figure}
    \centering
    \includegraphics[width=1\linewidth]{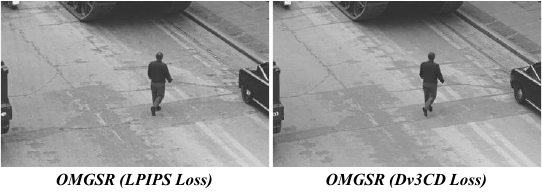}
    \caption{A artifact case in OMGSR-S with $\mathcal{L}_{\operatorname{LPIPS}}$ \vs $\mathcal{L}_{\operatorname{Dv3CD}}$.}
    \label{fig:artifact}
\end{figure}

Visually, as shown in~\cref{fig:artifact}, $\mathcal{L}_{\operatorname{LPIPS}}$ leads to noticeable artifacts, while $\mathcal{L}_{\operatorname{Dv3CD}}$ produces artifact-free results. This stems from the resolution limit of $\mathcal{L}_{\operatorname{LPIPS}}$, which is designed for inputs up to $224\times224$, which is far smaller than the $512\times512$ images typically used in Real-ISR. In contrast, $\mathcal{L}_{\operatorname{Dv3CD}}$ is built on DINOv3-ConvNeXt, which natively supports higher resolutions such as $512$ and beyond, enabling more stable and faithful image reconstruction.

\begin{figure*}
    \centering
    \includegraphics[width=1\linewidth]{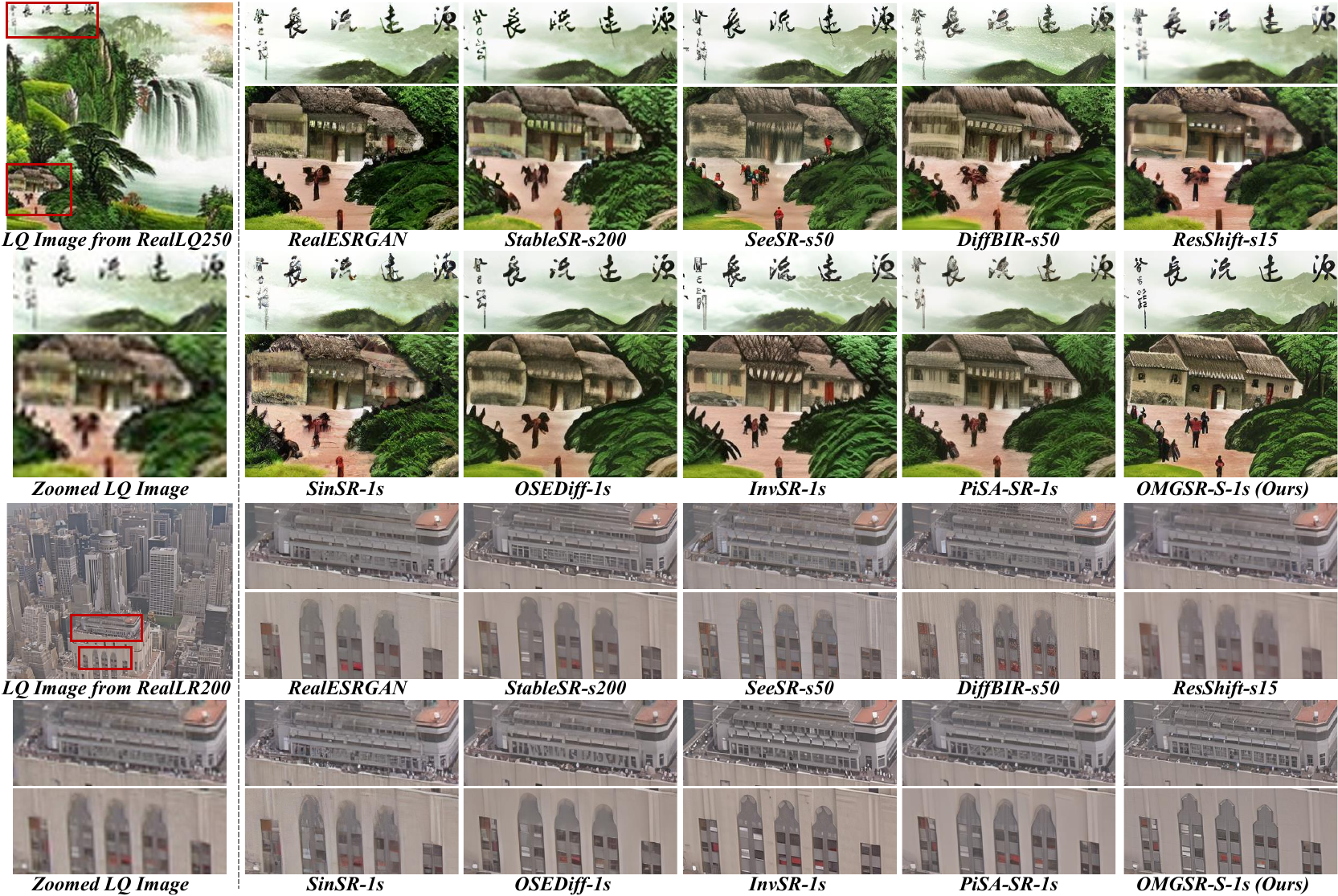}
    \caption{Qualitative comparisons with state-of-the-art multi-step and one-step models on two complex cases from RealLQ250 and RealLR200. Please zoom in for a detailed visualization. More qualitative comparisons can be found in the \textbf{supplementary materials}.}
    \label{fig:cmp}
\end{figure*}

\subsection{Comparison Results}
We compare OMGSR-S with the state-of-the-art SD-based models under the same foundation model size, including multi-step models (StableSR~\cite{stablesr}, SeeSR~\cite{seesr}, DiffBIR~\cite{diffbir}, ResShift~\cite{resshift}), and one-step models (OSEDiff~\cite{stablesr}, SinSR~\cite{sinsr}, InvSR~\cite{invsr}, PiSA-SR~\cite{pisasr}). 
We also include RealESRGAN~\cite{esrgan} as the baseline for reference. Here, we evaluate all the models on an RTX 4090.

\subsubsection{Quantitative Comparisons}
As summarized in \cref{tab:cmp}, our proposed OMGSR-S demonstrates highly competitive performance across all datasets and evaluation metrics. The quantitative results show that our method consistently achieves excellent performance, ranking first or second in most quality assessments. This consistent performance across diverse real-world benchmarks validates the effectiveness of our approach in handling various degradation patterns present in real images.
Considering that we employ $\mathcal{L}_{\operatorname{Dv3CD}}$ Loss instead of $\mathcal{L}_{\operatorname{LPIPS}}$, our method typically yields lower LPIPS metrics compared to models optimized with $\mathcal{L}_{\operatorname{LPIPS}}$, such as PiSA-SR and OSEDiff.
\subsubsection{Qualitative Comparisons}

Visual comparisons are presented in \cref{fig:cmp}. 
\textbf{First case:} In the first row, OSEDiff and OMGSR-S produce relatively higher textual quality compared to others, while OMGSR-S exhibits the best overall text clarity and structure. In the second row, RealESRGAN, SeeSR, ResShift, and SinSR struggle to generate the thatched cottage in the painting, whereas DiffBIR and OSEDiff roughly construct the basic form of the cottage. DiffBIR, InvSR, and PiSA-SR can generate basic thatched cottages. At the same time, our proposed OMGSR-S produces the best structural integrity and clarity in the generated cottage.
\textbf{Second case:} In the first row, StableSR, InvSR, and OMGSR reconstruct complex scenes more effectively than other models, with OMGSR-S demonstrating superior clarity and detail. In the second row, InvSR and OMGSR-S produce more plausible scene reconstructions, while OMGSR-S achieves even greater clarity.
Through these two complex scene cases, it can be observed that OMGSR-S excels in preserving finer details and structural coherence under challenging reconstruction scenarios.
\section{Conclusion}
\vspace{-2mm}
\textbf{Summary.} This work presents OMGSR, a universal GAN-based framework that employs a DDPM-based generative model as the generator and a DINOv3-ConvNeXt model with a multi-level discriminator as the discriminator. To effectively leverage generative priors, we first examine the role of mid-timestep injection in DDPM-based models and propose an SNR-based method to pre-compute an average optimal timestep. Furthermore, we introduce the Latent Representation Refinement (LRR) loss to enhance the VAE encoder by minimizing the latent representation gap. In addition, we design the DINOv3-ConvNeXt DISTS (Dv3CD) loss to improve structural perception, addressing the resolution mismatch between the model's supported input and the RealISR's training input. Within the OMGSR framework, we develop OMGSR-S (SD2.1-base), which achieves state-of-the-art performance across multiple metrics on four real-world datasets.
\textbf{Note.} The OMGSR framework is also compatible with flow matching~\cite{fm} models, such as FLUX.1‑dev~\cite{flux}. We also provide an analysis of flow matching within OMGSR and develop OMGSR‑F. Further details can be found in the \textbf{supplementary materials}.
{
    \small
    \bibliographystyle{ieeenat_fullname}
    \bibliography{main}
}

\clearpage
\setcounter{page}{1}
\maketitlesupplementary

\section{Extend Flow Matching to OMGSR}
\textbf{Flow Matching (FM).}
Unlike the DDPM scheduler, the FM scheduler maps $t$ into $\sigma_t \in (0,1)$ using a non-uniform spacing (\eg the shift strategies used in FLUX models). The pre-trained noisy latent distribution $\mathbf{z}_t$ of FM can be interpolated by $\mathbf{z}_0$ and $\epsilon$:
\begin{equation}
    \operatorname{FM}:~\mathbf{z}_t
    = \underbrace{(1 - \sigma_t) \cdot \mathbf{z}_H}_{\operatorname{Signal}} + \underbrace{\sigma_t \cdot \epsilon}_{\operatorname{Noise}}.
\end{equation}
In practical implementation, FM follows DDPM’s formulation for consistency, deviating from the original paper. The FM objective loss is 
\begin{equation}
    \mathcal{L}_{\operatorname{FM}}=\left\|\epsilon_\theta(\mathbf{z}_t, t, c) - (\epsilon - \mathbf{z}_H)\right\|^2_2,
\end{equation}
where $\theta$ is typically optimized by a DiT-based model $\epsilon_\theta$.

\noindent\textbf{Signal-to-Noise Ratio (SNR).}
FM-based models also involve a pre-trained noisy latent representation $\mathbf{z}_t$ at timestep $t$, which is a linear combination of Gaussian noise $\epsilon$ and the HQ image latent representation $\mathbf{z}_H$. 
The SNR of $\mathbf{z}_t$ is:
\begin{equation}
    \operatorname{FM}:~\texttt{SNR}(\mathbf{z}_t)=\frac{(1 - \sigma_t)^2  \cdot \mathbb{E}[\mathbf{z}_{H}^2]}{\sigma_t^2 \cdot \mathbb{E}[\epsilon^2]}=\frac{(1 - \sigma_t)^2 \cdot \mathbb{E}[\mathbf{z}_H^2]}{\sigma_t^2},
\end{equation}
where $\epsilon \sim \mathcal{N}(0, I)$ and $\mathbb{E}[\epsilon^2]=1$.

\noindent\textbf{Average Optimal Mid-timestep.}
Similar to \cref{sec:332}, the average optimal mid-timestep of the FM noisy latent representation can be pre-computed as:
\begin{equation}
    \operatorname{FM}:~t^\ast = \arg \min_t \frac{1}{N}\sum_i^N \left|\texttt{SNR}(\mathbf{z}_t^{(i)}) -\texttt{SNR}(\mathbf{z}_L^{(i)})\right|, 
    \label{eq19}
\end{equation}
\noindent\textbf{Note.}
When we precompute the $t^\ast$ of the FM-based generative model and obtain $\mathbf{z}_{t^\ast}$, the subsequent training of GAN is consistent with OMGSR-S.

\subsection{OMGSR-F (FLUX.1-dev)}
\begin{figure*}
    \centering
    \includegraphics[width=0.98\linewidth]{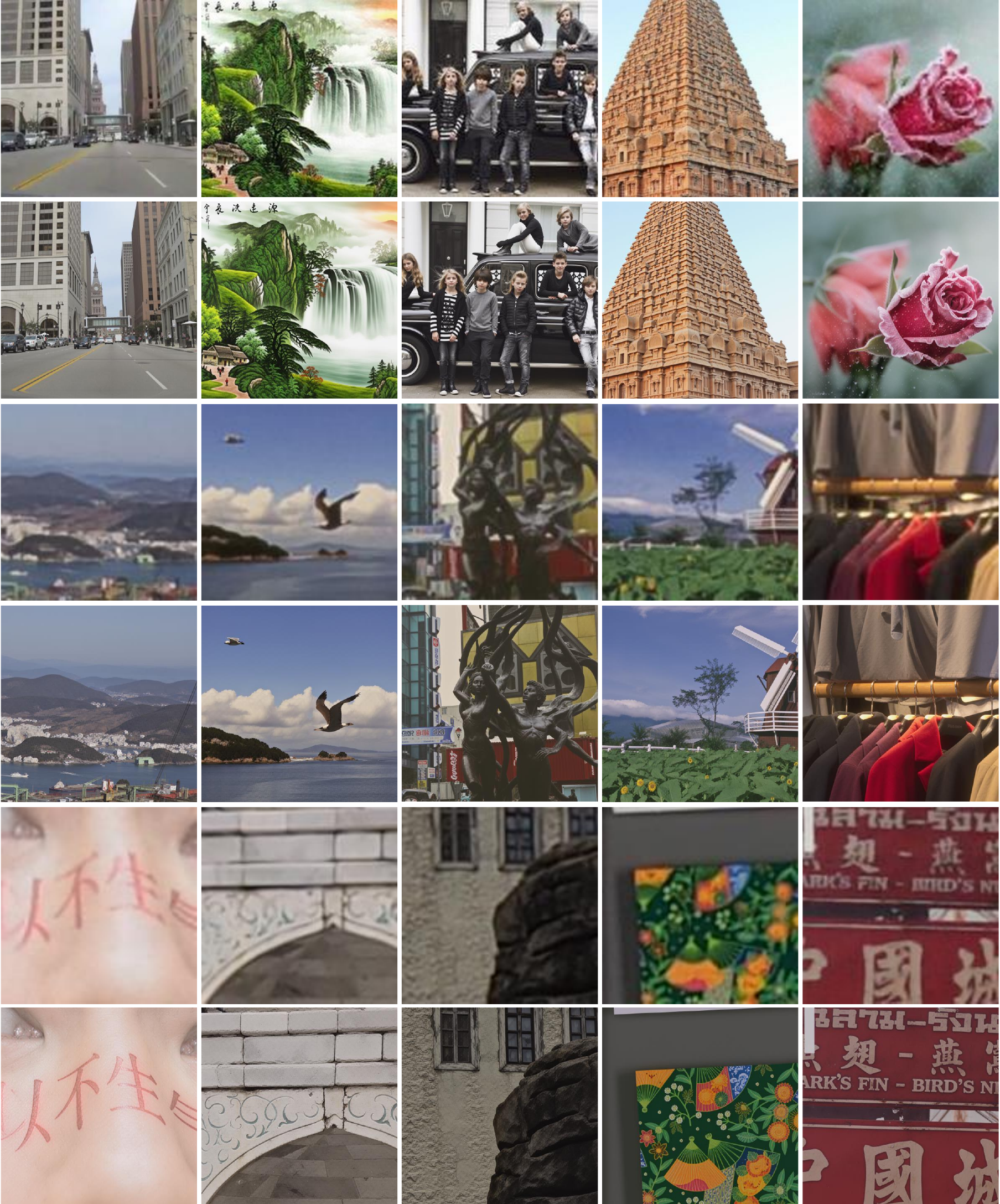}
    \caption{Visualization results of OMGSR-F. The first row is from the RealLQ250 dataset; the third row is from the RealSR dataset; the fifth line is from the DrealSR dataset. The second ($256\rightarrow 1024$), fourth ($128\rightarrow 1024$), and sixth ($128\rightarrow 1024$) lines are images generated by OMGSR-F. Please \textbf{zoom in}.}
    \label{fig:omgsr_f}
\end{figure*}

Based on FLUX.1-dev, we develop OMGSR-F, which successfully demonstrates the compatibility of our OMGSR with FM-based generative models. Note that OMGSR-F is trained at 1K-resolution, as the default resolution for FLUX.1-dev is 1K. We obtain the average optimal mid-timestep $t^{\ast}=244$ in FLUX.1-dev. Similar to OMGSR-S, OMGSR-F is trained on the LSDIR and FFHQ datasets. The training is conducted on dual H20 GPUs for $6000$ steps with a batch size of $1$, gradient accumulation of $4$, and a learning rate of $5\times10^{-5}$. \Cref{fig:omgsr_f} presents the performance of OMGSR-F on the RealLQ250, RealSR, and DrealSR datasets. From the figure, OMGSR-F exhibits strong global and detail generation capabilities.

\noindent\textbf{Note.} Our OMSR framework is theoretically compatible with DDPM- and FM-based generative models, such as SD1.5, SD2.1, SD3.5, FLUX.1-dev, and the latest Qwen-Image. For different SR tasks, we can choose different foundational models to adapt them. In this paper, we will not discuss it too much.

\section{Composite score on Timestep (OMGSR-S)}
As stated in \cref{sec:421}, we present the calculation of the composite score.
For the maximizing metrics (\ie CLIPIQA, CLIPIQA+, NIMA, LIQE, MUSIQ, MANIQA), the normalized score is calculated as 
\begin{equation}
    \mathcal{S}_{\operatorname{norm}} = \frac{\mathcal{S}-\mathcal{S}_{\operatorname{min}}}{\mathcal{S}_{\operatorname{max}}-\mathcal{S}_{\operatorname{min}}}.
\end{equation}
For the minimizing metrics (\ie LPIPS, FID, NIEQ), the normalized score is calculated as 
\begin{equation}
    \mathcal{S}_{\operatorname{norm}} = 1- \frac{\mathcal{S}-\mathcal{S}_{\operatorname{min}}}{\mathcal{S}_{\operatorname{max}}-\mathcal{S}_{\operatorname{min}}}.
\end{equation}
The composite score of each test dataset is defined: 
\begin{equation}
    \mathcal{S}_{\operatorname{comp}}^{\operatorname{Dataset}} = \frac{\sum \mathcal{S}_{\operatorname{norm}}^{\operatorname{Metict}}}{\operatorname{Metric~Count}}
\end{equation}
The average composite score of the four test datasets is defined: 
\begin{equation}
    \mathcal{S}_{\operatorname{comp}} = \frac{\mathcal{S}_{\operatorname{comp}}^{\operatorname{RealSR}} + \mathcal{S}_{\operatorname{comp}}^{\operatorname{DrealSR}} + \mathcal{S}_{\operatorname{comp}}^{\operatorname{RealLQ250}} + \mathcal{S}_{\operatorname{comp}}^{\operatorname{RealLR200}}}{4}
\end{equation}
We also provide the complete results in~\cref{tab:ab_t}. From the table, the best average composite score occurs at $273$ \ie $\mathcal{S}_{\operatorname{comp}} = (0.9793 + 0.9657 + 0.9712 + 0.9935) = 0.9774$.

\section{More Qualitative Comparisons (OMGSR-S)}
\begin{figure*}
    \centering
    \includegraphics[width=0.98\linewidth]{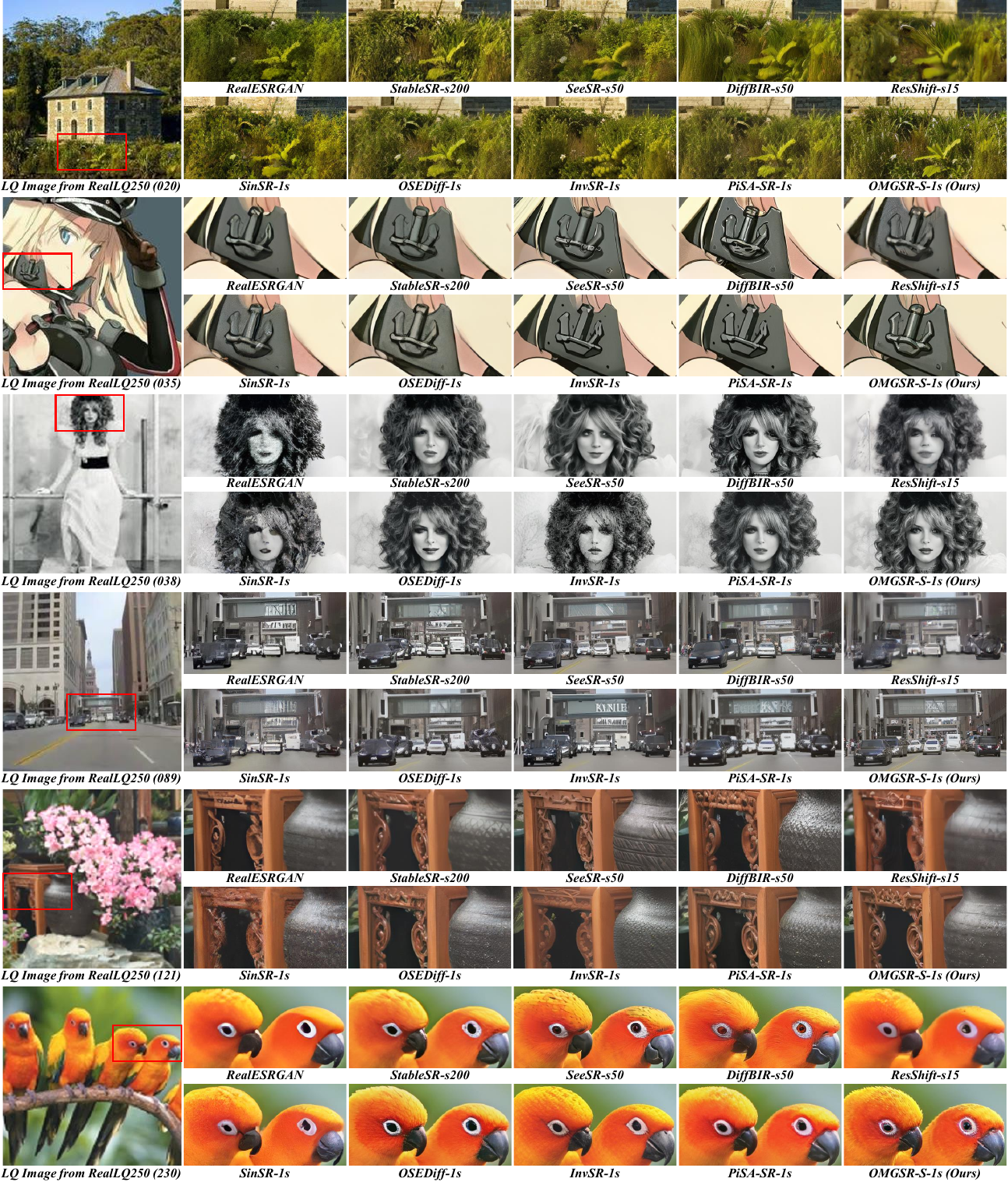}
    \caption{Qualitative comparisons with state-of-the-art multi-step and one-step models on RealLQ250. Please \textbf{zoom in}.}
    \label{fig:more_cmp}
\end{figure*}

\Cref{fig:more_cmp} shows additional qualitative comparisons, and it can be seen that OMGSR-S achieves excellent detail representation across multiple image scenes.

\begin{table*}[h]
  \caption{Ablation Study of OMGSR-S with different timesteps. We set the timestep $273$ as the baseline, with intervals of $100$. Note that we conduct experiments on OMGSR-S without the proposed LRR Loss to avoid any impact on the latent representation. We report a \textbf{Composite Score} based on $9$ metrics across four datasets. The best and second-best results of each metric are highlighted in bold and underlined, respectively.}
  \label{tab:ab_t}
  \resizebox{\textwidth}{!}{ 
  \centering
  \begin{tabular}{c|c|cccccccccccc}
    \toprule
    Datasets & Metrics & 999 & 973 & 873 & 773 & 673 & 573 & 473 & 373 & \cellcolor{gray!20}273 & 173 & 73 & 1 \\
    \midrule
    \multirow{10}{*}{RealSR} 
     & LPIPS$\downarrow$ & 0.4170 & 0.3702 & 0.3352 & 0.3003 & 0.2781 & 0.2868 & 0.2856 & 0.2788 & \cellcolor{gray!20}0.2815 & \textbf{0.2746} & \underline{0.2748} & 0.2776 \\
     & FID$\downarrow$ & 212.50 & 200.89 & 161.65 & 133.99 & 122.77 & 121.16 & 122.34 & \underline{117.69} & \cellcolor{gray!20}117.70 & \textbf{117.54} & 118.92 & 125.38 \\
     & CLIPIQA$\uparrow$ & 0.2630 & 0.3975 & \underline{0.6332} & 0.6236 & 0.6125 & 0.6254 & 0.6282 & 0.6265 & \cellcolor{gray!20}\textbf{0.6352} & 0.6184 & 0.6294 & 0.6170 \\
     & CLIPIQA+$\uparrow$ & 0.3239 & 0.3891 & 0.6448 & 0.6781 & 0.6706 & \textbf{0.6790} & 0.6766 & 0.6753 & \cellcolor{gray!20}\underline{0.6785} & 0.6737 & 0.6775 & 0.6735 \\
     & NIMA$\uparrow$ & 3.9255 & 4.5271 & \textbf{5.0184} & \underline{4.9240} & 4.8792 & 4.9174 & 4.9153 & 4.8934 & \cellcolor{gray!20}4.9030 & 4.8866 & 4.8638 & 4.8479 \\
     & NIQE$\downarrow$ & 7.9143 & 5.9296 & 5.9524 & 5.7239 & 5.7955 & \underline{5.6473} & \textbf{5.6115} & 5.7276 & \cellcolor{gray!20}5.6487 & 5.7333 & 5.6939 & 5.6876 \\
     & LIQE$\uparrow$ & 1.1054 & 2.1472 & 4.0819 & 4.1027 & 4.0200 & \underline{4.1557} & 4.1047 & 4.1279 & \cellcolor{gray!20}4.1373 & 4.1386 & \textbf{4.1745} & 4.1151 \\
     & MUSIQ$\uparrow$ & 37.7757 & 53.6943 & 68.2292 & 68.5141 & 68.1226 & 68.1545 & 68.2329 & \textbf{68.5897} & \cellcolor{gray!20}\underline{68.5648} & 68.1914 & 68.2788 & 68.1661 \\
     & MANIQA$\uparrow$ & 0.3534 & 0.4158 & 0.6230 & 0.6610 & 0.6697 & \underline{0.6721} & 0.6706 & 0.6712 & \cellcolor{gray!20}\textbf{0.6756} & 0.6711 & 0.6703 & 0.6500 \\
     & Composite Score$\uparrow$ & 0.0 & 0.3842 & 0.8506 & 0.9341 & 0.9494 & 0.9679 & 0.9669 & 0.9713 & \cellcolor{gray!20}\textbf{0.9793} & 0.9698 & \underline{0.9734} & 0.9478 \\
    \midrule
    \multirow{10}{*}{DrealSR} 
     & LPIPS$\downarrow$ & 0.3760 & 0.3671 & 0.3430 & 0.3122 & \textbf{0.2931} & 0.3047 & 0.3076 & 0.3046 & \cellcolor{gray!20}0.3027 & \underline{0.2998} & 0.3025 & 0.3018 \\
     & FID$\downarrow$ & 212.27 & 202.16 & 183.24 & 150.05 & \underline{134.97} & 138.40 & 135.53 & 138.61 & \cellcolor{gray!20}\textbf{132.77} & 137.90 & 139.04 & 145.07 \\
     & CLIPIQA$\uparrow$ & 0.3044 & 0.4161 & 0.5974 & 0.6346 & 0.6245 & 0.6539 & 0.6565 & \underline{0.6584} & \cellcolor{gray!20}0.6552 & 0.6491 & \textbf{0.6640} & 0.6569 \\
     & CLIPIQA+$\uparrow$ & 0.3287 & 0.4042 & 0.5852 & 0.6448 & 0.6381 & 0.6659 & 0.6505 & 0.6610 & \cellcolor{gray!20}0.6650 & 0.6595 & \underline{0.6661} & \textbf{0.6672} \\
     & NIMA$\uparrow$ & 3.7182 & 4.3704 & 4.6844 & 4.6728 & 4.6659 & 4.7237 & 4.7080 & 4.7061 & \cellcolor{gray!20}\textbf{4.7416} & \underline{4.7303} & 4.7125 & 4.6954 \\
     & NIQE$\downarrow$ & 8.7497 & \textbf{6.2222} & 6.5924 & 6.6827 & 6.5815 & 6.6154 & 6.4873 & 6.5317 & \cellcolor{gray!20}6.5184 & 6.7222 & 6.5704 & \underline{6.3530} \\
     & LIQE$\uparrow$ & 1.1721 & 2.5310 & 3.7494 & 3.8657 & 3.8151 & 3.8832 & 3.9936 & 3.9849 & \cellcolor{gray!20}3.9885 & 4.0250 & \textbf{4.0618} & \underline{4.0270} \\
     & MUSIQ$\uparrow$ & 36.8816 & 52.5906 & 62.4311 & 63.6773 & 63.2950 & 64.9613 & 65.1226 & 65.0771 & \cellcolor{gray!20}\underline{65.3381} & 65.0661 & \textbf{65.6084} & 65.2814 \\
     & MANIQA$\uparrow$ & 0.3679 & 0.4163 & 0.5700 & 0.6023 & 0.6160 & 0.6210 & 0.6234 & 0.6244 & \cellcolor{gray!20}0.6259 & \underline{0.6277} & \textbf{0.6285} & 0.6039 \\
     & Composite Score$\uparrow$ & 0.0 & 0.4009 & 0.7434 & 0.8799 & 0.9274 & 0.9412 & 0.9469 & 0.9484 & \cellcolor{gray!20}\textbf{0.9657} & 0.9497 & \underline{0.9598} & 0.9451 \\
    \midrule
    \multirow{8}{*}{RealLQ250} 
     & CLIPIQA$\uparrow$ & 0.2994 & 0.4472 & 0.5647 & 0.6503 & 0.6570 & 0.6945 & 0.6931 & 0.6938 & \cellcolor{gray!20}\textbf{0.6976} & \underline{0.6952} & 0.6946 & 0.6908 \\
     & CLIPIQA+$\uparrow$ & 0.3716 & 0.4657 & 0.5949 & 0.6800 & 0.6836 & 0.7052 & 0.7005 & 0.7017 & \cellcolor{gray!20}\underline{0.7065} & \textbf{0.7073} & 0.7012 & 0.7025 \\
     & NIMA$\uparrow$ & 4.5873 & \textbf{5.2518} & \underline{5.1600} & 5.0815 & 5.1142 & 5.1088 & 5.1168 & 5.0935 & \cellcolor{gray!20}5.1490 & 5.1254 & 5.1224 & 5.1324 \\
     & NIQE$\downarrow$ & 7.2817 & 4.4344 & 4.4566 & 4.1618 & \textbf{3.9151} & 4.1169 & 4.0448 & 4.1001 & \cellcolor{gray!20}\underline{4.0240} & 4.0927 & 4.1152 & 4.0679 \\
     & LIQE$\uparrow$ & 1.1207 & 2.2094 & 3.1658 & 3.7760 & 3.7987 & \underline{3.9796} & 3.9433 & 3.9234 & \cellcolor{gray!20}\textbf{3.9954} & 3.8639 & 3.8418 & 3.8582 \\
     & MUSIQ$\uparrow$ & 43.9835 & 54.3022 & 63.0905 & 67.6347 & 68.3741 & 69.5356 & 69.6991 & \underline{69.7241} & \cellcolor{gray!20}\textbf{69.8671} & 69.1404 & 69.4704 & 69.2748 \\
     & MANIQA$\uparrow$ & 0.3304 & 0.3749 & 0.5446 & 0.6084 & 0.6175 & 0.6255 & \textbf{0.6339} & 0.6302 & \cellcolor{gray!20}0.6302 & \underline{0.6308} & 0.6290 & 0.6139 \\
     & Composite Score$\uparrow$ & 0.0 & 0.4887 & 0.7411 & 0.8891 & 0.9200 & 0.9521 & \underline{0.9574} & 0.9483 & \cellcolor{gray!20}\textbf{0.9712} & 0.9524 & 0.9479 & 0.9439 \\
    \midrule
    \multirow{8}{*}{RealLQ200} 
     & CLIPIQA$\uparrow$ & 0.2996 & 0.4508 & 0.5997 & 0.6550 & 0.6641 & 0.6962 & 0.6944 & 0.6920 & \cellcolor{gray!20}\textbf{0.7016} & 0.6971 & \underline{0.7001} & 0.6854 \\
     & CLIPIQA+$\uparrow$ & 0.3740 & 0.4762 & 0.6282 & 0.6953 & 0.7017 & 0.7197 & \underline{0.7205} & 0.7198 & \cellcolor{gray!20}\textbf{0.7211} & 0.7202 & 0.7201 & 0.7162 \\
     & NIMA$\uparrow$ & 4.4772 & 5.0967 & 5.3409 & 5.2591 & 5.2736 & 5.2953 & 5.3173 & 5.3153 & \cellcolor{gray!20}\textbf{5.3617} & \underline{5.3425} & 5.3363 & 5.3243 \\
     & NIQE$\downarrow$ & 7.2073 & 4.4231 & 4.5819 & 4.2892 & 4.2503 & 4.1507 & 4.1882 & 4.1302 & \cellcolor{gray!20}\underline{4.0731} & 4.1353 & 4.1780 & \textbf{3.9494} \\
     & LIQE$\uparrow$ & 1.1756 & 2.2239 & 3.5141 & 3.9733 & 4.0361 & 4.1637 & 4.1497 & \underline{4.1688} & \cellcolor{gray!20}\textbf{4.1901} & 4.1684 & 4.1513 & 4.1417 \\
     & MUSIQ$\uparrow$ & 43.1193 & 55.2588 & 63.9919 & 67.3852 & 68.5206 & 69.6565 & 69.7687 & 69.7503 & \cellcolor{gray!20}\textbf{69.8960} & 69.1132 & 69.4267 & \underline{69.8007} \\
     & MANIQA$\uparrow$ & 0.3448 & 0.4000 & 0.5859 & 0.6368 & 0.6477 & 0.6527 & 0.6537 & 0.6567 & \cellcolor{gray!20}\underline{0.6571} & \textbf{0.6594} & 0.6529 & 0.6449 \\
     & Composite Score$\uparrow$ & 0.0 & 0.4574 & 0.7975 & 0.9074 & 0.9313 & 0.9724 & 0.9744 & 0.9776 & \cellcolor{gray!20}\textbf{0.9935} & \underline{0.9816} & 0.9776 & 0.9768 \\
     \midrule
     & \makecell{Average \\ Composite Score}$\uparrow$ & 0.0 & 0.4328 & 0.7831 & 0.9026 & 0.932 & 0.9584 & 0.9614 & 0.9614 & \cellcolor{gray!20}\textbf{0.9774} & \underline{0.9634} & 0.9647 & 0.9534 \\
    \bottomrule
  \end{tabular}
  }
\end{table*}

\end{document}